\documentclass[11pt]{article}

\usepackage[utf8]{inputenc}
\usepackage{deauthor}
\usepackage{times,graphicx}
\usepackage{url}

\usepackage{todonotes}
\usepackage{pifont}

\usepackage{multirow}
\usepackage{booktabs}
\usepackage{caption,subcaption}
\usepackage{graphicx}
\usepackage{amsmath}
\usepackage[numbers]{natbib}

\title{Top Ten Challenges Towards Agentic Neural Graph Databases}

\author{Jiaxin Bai\thanks{~~Equal Contribution} $^{\dagger}$
\hspace{0.5em} Zihao Wang\footnotemark[1] $^{\dagger}$
\hspace{0.5em} Yukun Zhou$^{\dagger}$
\hspace{0.5em} Hang Yin$^{\ddagger}$
\hspace{0.5em} Weizhi Fei$^{\ddagger}$
\hspace{0.5em} Qi Hu$^{\dagger}$
\hspace{0.5em} Zheye Deng$^{\dagger}$ \\
\hspace{0.5em} Jiayang Cheng$^{\dagger}$ 
\hspace{0.5em} Tianshi Zheng$^{\dagger}$
\hspace{0.5em} Hong Ting Tsang$^{\dagger}$
\hspace{0.5em} Yisen Gao$^{\vee}$
\hspace{0.5em} Zhongwei Xie$^{\bot}$
\hspace{0.5em} Yufei Li$^{\top}$\\
\hspace{0.5em} Lixin Fan$^{\mathparagraph}$
\hspace{0.5em} Binhang Yuan$^{\dagger}$
\hspace{0.5em} Wei Wang$^{\dagger}$
\hspace{0.5em} Lei Chen$^{\dagger}$ 
\hspace{0.5em} Xiaofang Zhou$^{\dagger}$ 
\hspace{0.5em} Yangqiu Song$^{\dagger}$ \vspace{.5em}\\
$^{\dagger}$ Department of Computer Science and Engineering, HKUST, Hong Kong, China\\
$^{\ddagger}$ Department of Mathematical Sciences, Tsinghua University, Beijing, China \hspace{0.5em} $^{\mathparagraph}$ AI Group, WeBank\\
$^{\vee}$ Institute of Artificial Intelligence, Beihang University \hspace{0.5em}  $^{\bot}$ Wuhan University  \hspace{0.5em}$^{\top}$ Sichuan University \\
 \texttt{\small\{jbai,zwanggc,yzhoufw,qhuaf,zdengah,jchengaj,tzhengad,httsangaj\}@cse.ust.hk}\\ 
 \texttt{\small\{biyuan, weiwa, leichen, zxf, yqsong\}@cse.ust.hk}\\
 \texttt{\small  lixinfan@webank.com} 
 \texttt{\small \{h-yin20, fwz22\}@mails.tsinghua.edu.cn}\\
\texttt{\small  yisengao@buaa.edu.cn} 
\hspace{0.5em}\texttt{\small zhongwei.xie@whu.edu.cn}
\hspace{0.5em}\texttt{\small  evangeline@stu.scu.edu.cn} 
}

\usepackage{paralist}

\begin{document}

\maketitle

\begin{abstract}
Graph databases (GDBs) like Neo4j and TigerGraph excel at handling interconnected data but lack advanced inference capabilities. Neural Graph Databases (NGDBs) address this by integrating Graph Neural Networks (GNNs) for predictive analysis and reasoning over incomplete or noisy data. However, NGDBs rely on predefined queries and lack autonomy and adaptability.
This paper introduces Agentic Neural Graph Databases (Agentic NGDBs), which extend NGDBs with three core functionalities: autonomous query construction, neural query execution, and continuous learning. We identify ten key challenges in realizing Agentic NGDBs: semantic unit representation, abductive reasoning, scalable query execution, and integration with foundation models like large language models (LLMs). By addressing these challenges, Agentic NGDBs can enable intelligent, self-improving systems for modern data-driven applications, paving the way for adaptable and autonomous data management solutions.
\end{abstract}

\section{Introduction}

Graph databases like Neo4j~\cite{NGDB:miller2013graph}, TigerGraph~\cite{NGDB:Deutsch2019TigerGraphAN}, and Azure Cosmos DB are useful tools for representing and querying interconnected data using nodes and edges. These databases are adept at handling the complex relationships inherent in graph-structured data, providing efficient mechanisms for storage and retrieval. 

A Neural Graph Database (NGDB), as introduced in~\cite{NGDB:DBLP:conf/log/BestaISODPCH22}, represents a system architecture that merges the predictive capabilities of Graph Neural Networks (GNNs) with the rich data representation features of graph databases (GDBs). 
NGDBs enhance graph databases by leveraging GNNs for advanced machine-learning tasks while preserving and utilizing the information embedded within the graph data model.

However, methodologies for conducting inferences within this latent neural space are yet to be thoroughly explored. To address this gap, the integration of neural execution engines on top of neural graph storage has been proposed~\cite{NGDB:ren2023neural}. 
By utilizing neural embeddings and neural networks, NGDBs enhance their ability to perform complex reasoning and more effectively infer hidden relationships, which are the capabilities that traditional graph databases lack. 
This fusion of symbolic graph representations with neural computation paves the way for more intelligent and adaptable data management systems to address contemporary applications' diverse demands. 
The process of ``neuralization'' is particularly beneficial for inferring missing information within the underlying graph data model, enriching the database with additional knowledge.

From a broader perspective, the principles of data management systems revolve around efficiently storing, retrieving, and managing data while providing a layer of abstraction to users. These systems aim to handle large volumes of data and complex operations, concealing the underlying complexities from end-users. Motivated by this principle, we propose the concept of \textbf{Agentic Neural Graph Databases (Agentic NGDBs)}, extending neuralization to further automate data and data management processes. 
Here, we summarize the challenges regarding the Agentic NGDB from the following three perspectives interface, learning, and system: 

\begin{itemize}
    \item \textbf{Interface}: The Agentic NGDB should automatically construct appropriate queries that generate useful answers for a given task in a specific context.
    \item \textbf{Learning and Inference}: Agentic NGDB should leverage neural networks to execute queries and derive meaningful answers as neural network predictions, even when the underlying data model is incomplete.
    \item \textbf{System}: The Agentic NGDB should remain compatible with existing graph databases, supporting most standard GDB operators. Additionally, it should function as an adaptor for foundation models, enhancing knowledge and reasoning capabilities. Furthermore, it must actively learn by constructing and executing appropriate \texttt{CREATE}, \texttt{UPDATE}, or \texttt{DELETE} queries in a given context.
\end{itemize}

\begin{figure}[t]
\centering
    \includegraphics[width=\linewidth]{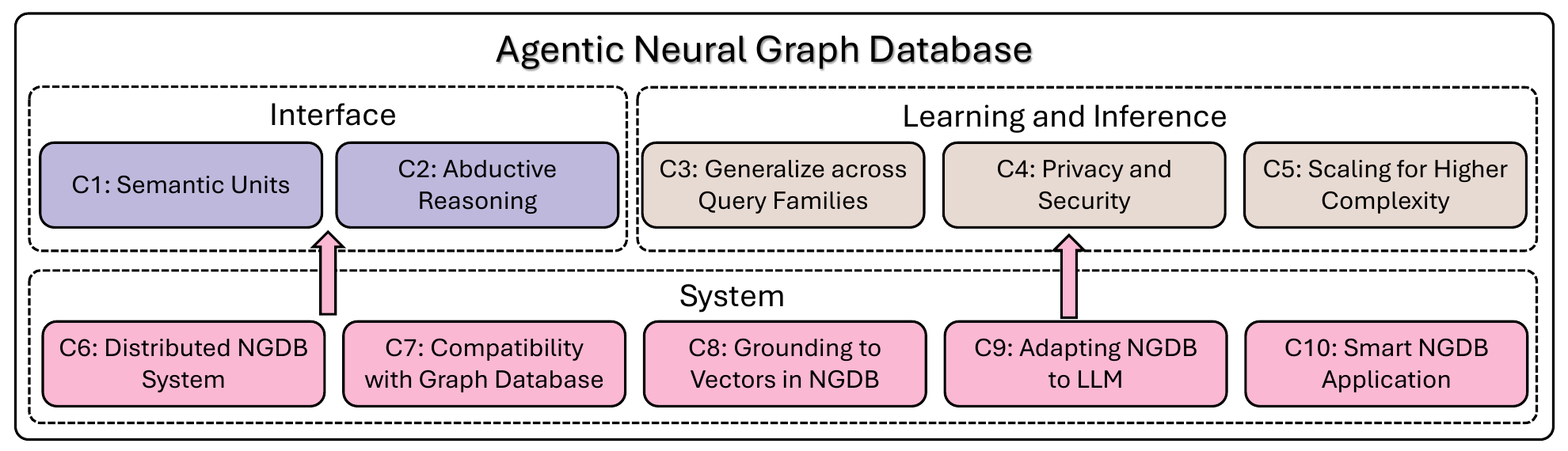}
    \vspace{-0.5cm}
  \caption{The top ten challenges in achieving Agentic NGDB. Its three perspectives include interface, learning, and system.}
  \label{fig:ten_challenges}
 
\end{figure}

There are significant challenges to achieving each of these aspects, as illustrated in Figure~\ref{fig:ten_challenges}. We identify the most critical challenges for realizing these functionalities based on recent progress in the research community on logical query answering and logical hypothesis generation for relational graphs.

\paragraph{Interface}
The first significant challenge in the \textbf{Interface} component is addressing \textit{fundamental semantic units} (\textbf{Challenge 1}) within the neural graph database's query and data model. Semantic units refer to the data types associated with nodes and edges, such as atomic IDs, text strings (e.g., entities and events), numbers, and dates. Constructing queries that effectively handle these diverse semantic units presents a significant obstacle. Beyond managing individual semantic units, another critical challenge lies in connecting these units to construct more complex queries.
The \textbf{Interface} component also requires advanced \textit{abductive reasoning capabilities} (\textbf{Challenge 2}). In Agentic NGDBs, abductive reasoning refers to identifying the optimal NGDB query that best explains or supports a specific task in a given context. This capability ensures that the database can adaptively generate meaningful, task-relevant queries. The generated queries can then be executed symbolically by a graph database or through neural execution within the system.

\paragraph{Learning and Inference}
Neural query execution is the core functionality of traditional NGDBs, referring to the ability to perform tasks or actions according to a predefined plan or strategy within the neural space~\cite{NGDB:russell2016artificial} by learning and inferences. However, several practical challenges remain in this area.
One major challenge is enhancing \textit{inference capabilities for better generalization} (\textbf{Challenge 3}) across query families. This involves ensuring that NGDB systems can effectively handle and generalize across diverse query structures and types, even when presented with novel or complex combinations of queries. Another critical hurdle involves maintaining \textit{data privacy and security} (\textbf{Challenge 4}). Due to their inherent vulnerability to extracting latent knowledge, NGDBs must safeguard sensitive information against advanced inference attacks, particularly in neural models. Robust privacy mechanisms are essential for building trust and ensuring security in NGDB applications.
The \textit{scaling laws of neural query execution} (\textbf{Challenge 5}) must be explored. Scaling laws in NGDBs describe how the system's performance improves as key factors, such as the number of model parameters, the size of the training dataset, and training costs, are increased. This concept is rooted in neural scaling laws observed in deep learning, where larger models generally lead to better performance, albeit at higher computational costs.

\paragraph{System}
The \textbf{System} component focuses on building a system on top of the learning and inference algorithms that can ensure continuous learning and adaptation within Agentic NGDBs.
Efficiently processing and managing large-scale data while maintaining high performance becomes especially critical when dealing with massive datasets. This challenge is further amplified in distributed NGDB architectures, where optimizing query performance under read-intensive workloads and dynamically fluctuating demands is necessary. 
Ensuring elastic scalability and developing NGDBs that operate effectively as \textit{distributed systems} (\textbf{Challenge 6}) are key to achieving these goals.
These systems must be capable of improving themselves by writing and executing \texttt{CREATE}, \texttt{UPDATE}, and \texttt{DELETE} clauses or performing model editing directly within the neural latent space.
The first aspect of this functionality involves ensuring \textit{compatibility with graph database models} (\textbf{Challenge 7}). The fundamental CRUD (CREATE, READ, UPDATE, DELETE) actions are essential for managing and modifying persistent data elements in traditional graph databases. Seamlessly integrating these actions into NGDBs is necessary for enabling effective self-improvement.
The second aspect involves \textit{grounding vectors within NGDBs} (\textbf{Challenge 8}). For effective learning and adaptation, the system must accurately identify the locations of relevant knowledge and understand how reasoning is conducted within the latent neural space. Proper grounding ensures that modifications and updates align with the underlying knowledge representation.
Moreover, the Agentic NGDB must be capable of \textit{integrating with foundation models}, such as large language models (LLMs), to enhance its reasoning and knowledge capabilities (\textbf{Challenge 9}). The NGDB can provide more reliable and contextually accurate results for various tasks by leveraging foundation models' advanced natural language understanding and reasoning capabilities. 
Finally, we discuss the challenge of developing Smart Neural Graph Databases (NGDB) applications (\textbf{Challenge 10}), particularly Agentic NGDB. The challenges lie in creating systems that leverage their advanced functionalities across diverse applications.

\paragraph{} Agentic NGDBs extend the capabilities of traditional NGDBs by incorporating autonomy, active learning, and adaptability. While NGDBs enhance graph databases by integrating Graph Neural Networks (GNNs) to perform advanced inference and reasoning and handle incomplete or noisy data, they rely heavily on predefined tasks and human-defined queries. In contrast, Agentic NGDBs introduce three core functionalities: automatic query construction tailored to specific tasks and contexts, neural query execution for predictive analysis, and continuous learning and adaptation through active updates to the knowledge base. In the following sections, we will individually discuss each identified challenge.

\vspace{-.5em}
\section{Challenge 1: Semantic Units}
\vspace{-.5em}

The NGDB primarily relies on relational graphs, where nodes and relations are the basic semantic units. Incorporating diverse semantic units, such as numbers and events, introduces complexity due to their intrinsic relationships. For example, numbers involve algebraic operations (e.g., addition, subtraction), while events involve temporal and causal relations. Addressing these complexities requires reasoning engines that can learn and process such relationships effectively.

Number literals (e.g., age, height) are critical for filtering and querying within NGDBs. Prior work includes methods like KBLRN \cite{NGDB:garcia2017kblrn}, KR-EAR \cite{NGDB:lin2016knowledge}, and LitCQD \cite{NGDB:DBLP:conf/pkdd/DemirWLNH23}, which improve reasoning by integrating numeric constraints into queries.
Despite these advancements, challenges remain. These include developing advanced numerical operations and integrating neural-symbolic systems into NGDBs while ensuring compatibility with symbolic solvers for faithful reasoning.
Existing approaches focus on entity-centric knowledge graphs, but event-centric knowledge graphs (EVKGs) like ATOMIC \cite{NGDB:DBLP:conf/aaai/SapBABLRRSC19} and ASER \cite{NGDB:DBLP:conf/www/ZhangLPSL20} emphasize relationships between events (e.g., temporal and causal relations). Reasoning on EVKGs involves determining event occurrences and their sequences, which introduces unique challenges compared to entity-centric KGs.
Recent work extends traditional reasoning by integrating temporal and occurrence constraints\cite{NGDB:bai2023complexqueryansweringeventuality}.

% \subsection{Theory of Mind related}

Moreover, {beliefs, desires, and intentions (BDI)} represent higher-level, abstract semantic units extending beyond simple entity-attribute relationships and eventualities. 
These elements are crucial for modeling human-like reasoning, decision-making, and behavior prediction. {Beliefs} refer to what an agent (human or system) assumes or holds to be true about the world. These can include factual statements like \textit{It is raining outside} and subjective perspectives like \textit{This movie is great}. In KGs, beliefs are often represented as knowledge nodes or statements that may vary across agents or contexts, allowing for personalization or multi-agent reasoning. 
{Intentions} represent the goals or purposes behind an agent's actions or decisions and as a bridge between beliefs and actions. 
Intentions are often implicit and must be inferred from user behavior or contextual information. KGs are typically modeled as motivational nodes or goals that guide reasoning about why an agent performs specific actions. For instance, \textit{PersonX intends [to buy a gift for a friend]}, which could explain why \textit{PersonX searches for [gift shops nearby]}. On the other hand, {desires} represent an agent's wants, preferences, or needs, which may not always lead to concrete actions unless accompanied by intention. In knowledge graphs, desires are commonly expressed as preferences or motivational entities that influence behavior, such as \textit{PersonX desires [to eat ice cream]}. These three elements allow knowledge graphs to capture human motivations more comprehensively.
These concepts are closely connected to the Theory of Mind (ToM), which refers to the ability to understand that other agents (humans, machines, etc.) possess their own beliefs, desires, and intentions that may differ from one’s own. In the context of knowledge graphs, the Theory of Mind enhances reasoning about multi-agent knowledge by enabling the understanding of diverse perspectives. 
Theory of Mind also enables the inference of motivations by reasoning about the interplay between beliefs, desires, and intentions. 

Integrating BDI and ToM in Agentic NGDB has practical applications across various domains. 
In e-commerce, systems like FolkScope \cite{NGDB:yu-etal-2023-folkscope},  COSMO \cite{NGDB:Yu2024}, and RIG \cite{NGDB:bai2024intention} are the knowledge graphs that leverage BDI to model user behavior, enabling personalized recommendations by linking user actions (e.g., purchases) with inferred desires and intentions. In commonsense reasoning, resources like ATOMIC use BDI to represent cause-effect relationships, allowing systems to reason about potential outcomes of actions. Multi-agent systems benefit from BDI-enhanced KGs by enabling cooperative and competitive interactions that account for multiple agents' goals, beliefs, and desires. Additionally, in natural language understanding, BDI helps interpret user intent in queries, conversations, and social media posts by associating semantic meanings with inferred motivations. We still need systematic storing and inference with these intention knowledge graphs.

\vspace{-.5em}
\section{Challenge 2: Abductive Reasoning with NGDB}
\vspace{-.5em}
Abductive reasoning, the process of inferring the most plausible explanations for observations, is a fundamental aspect of human cognition and artificial intelligence. In the context of knowledge graphs (KGs), abductive reasoning generates hypotheses to explain observations (entity sets) by leveraging structured relationships and entities. Complex Logical Query Answering (CLQA) has further advanced abductive reasoning by enabling multi-hop logical inferences over large, incomplete graphs. Neural Graph Databases (NGDBs) build on these advancements, offering a more flexible and robust framework for abductive reasoning.

Early methods for abductive reasoning in KGs relied on supervised learning and search-based techniques. Generative models, such as transformer-based architectures, were used to produce logical hypotheses. For example, \cite{NGDB:Bai2023AdvancingAR} proposed a supervised generative model trained on datasets like FB15k-237 and WN18RR, which excelled in structural fidelity but struggled to generalize to unseen observations due to the limitations of supervised objectives.
To address these limitations, reinforcement learning (RL) techniques were introduced. Reinforcement Learning from Knowledge Graph feedback (RLF-KG) employed proximal policy optimization (PPO) to generate hypotheses aligned with observed evidence. This approach improved explanatory power and generalizability, achieving significant gains in metrics like Jaccard similarity and Smatch scores across multiple datasets.
NGDBs extend these methods by embedding knowledge graph data in a latent space, enabling flexible query processing and hypothesis generation. By leveraging latent embeddings, NGDBs can infer missing information and generate hypotheses for complex logical queries, even on incomplete graphs, outperforming traditional graph databases.
NGDBs represent a significant step forward in abductive reasoning, synthesizing the strengths of CLQA and advanced generative models. However, several challenges must be addressed:
\begin{itemize}
    \item \textbf{More Generalized Observation}
In the current definition of abductive reasoning, the definition of the observations is a set of entities. 
However, observation can be further generalized to a context, for example, a conversation history in the conversational recommendation task setting, or a structured shopping session.
\item \textbf{More Complex Structured Hypotheses}
Existing abductive reasoning models on KGs primarily focus on conjunctive tree-formed queries. NGDBs, with their increased query expressiveness, require hypothesis generation models capable of handling more complex structured observations. For instance, hypotheses should accommodate EFO$_k$ (existential first-order logic) and cyclic queries, expanding beyond the limitations of earlier models.
\item \textbf{Graph-Based Hypothesis Generation Models}
Traditional sequence-based models struggle to capture the structural complexity of logical hypotheses, which are fundamentally query graphs. These graphs exhibit features like permutation invariance of logical operators, requiring models explicitly designed to generate graph-structured hypotheses.
\item \textbf{NGDB as a Reward Model for Reinforcement Learning}
Previous RL-based methods, such as \cite{NGDB:Bai2023AdvancingAR}, relied on symbolic execution results from knowledge graphs to provide reward signals during hypothesis generation. However, these reward signals suffer from the incompleteness inherent to the open-world assumption. NGDBs can address this issue by serving as a more robust reward model, leveraging their latent embeddings and flexible query capabilities to improve hypothesis generation.
\end{itemize}

\vspace{-.5em}
\section{Challenge 3: Generalization across Query Families}
\vspace{-.5em}

Introducing neural modules in graph databases enables the generalization to the \textit{knowledge} in databases. However, the development of neural modules is always entangled with their targeted query families, thus naturally biased toward them due to their inductive biases, emphasizing the challenge of generalizing towards different \textit{query types}. Compared to classic database algorithms that support an entire query family as long as it is formally defined, neural modules still suffer a loss in performance for generalization even when the query family is fixed~\cite{NGDB:wang_benchmarking_2021}. Readers are also referred to related surveys~\cite{NGDB:wang2022logical,NGDB:liu2024neural}.

\subsection{Different Query Families and Their Neural Modules}

\paragraph{Tree-formed Queries and Compositional Generalizability.}
The tree-formed query is a collective term that describes the whole query family that can be recursively defined in a tree structure, in which logical connectives and variables are carefully organized so that set operations can formally derive the answers~\cite{NGDB:wang_benchmarking_2021}, the set operations include set projection~\cite{NGDB:hamilton_embedding_2018}, intersection, union~\cite{NGDB:ren_query2box_2020}, complement\cite{NGDB:ren_query2box_2020} and set difference~\cite{NGDB:liu_neural-answering_2021}. To tackle such kinds of queries, a line of research is known as {query embeddings}, where sets are modeled as embeddings, and set operations mentioned above are modeled directly by neural modules~\cite{NGDB:ren_beta_2020,NGDB:liu_neural-answering_2021,NGDB:wang2023wasserstein}. The set operations composition allows the models to generalize the entire tree-form query family. This connection between model design and query family is termed the compositional generalizability~\cite{NGDB:wang_benchmarking_2021,NGDB:yin2024meta}, and the performance drop with the increasing of compositional levels is still universally observed and remains a challenge to address.

\paragraph{EFO-1 Queries and Query Graph.}
It is shown that tree-formed query family is constrained by certain assumptions and fails to represent the whole family of Existential First Order queries with one free variable (EFO-1 query) such as cyclic query~\cite{NGDB:yin_rethinking_2023}. To handle new graph-theoretic features which cannot be represented in tree-formed queries. One commonly adopted technique for EFO-1 queries is the DNF normal form or the UCQ query-solving strategy~\cite{NGDB:ren_query2box_2020,NGDB:wang2023logical}, which solves the conjunctive query first and then takes the union of the answer set of each conjunctive query. A query graph~\cite{NGDB:wang2023logical} can naturally describe each conjunctive query. This formulation motivates graph-related search methods~\cite{NGDB:yin_rethinking_2023} or graph neural networks~\cite{NGDB:wang2023logical}.

\paragraph{More Advanced Query Types.}
More advanced query families still exist, though the development of corresponding neural models on these topics is insufficient at the current stage. Thus, we discuss some of the challenges we might face in pursuit of more advanced queries in NGDB from the following aspects
\textbf{(i) Multi-arity predicates:}
The first challenge we may encounter is when the knowledge databases are constructed by $(n+1)$-ary tuples, the relation corresponds to $n$-ary predicate and a graph becomes a hypergraph~\cite{NGDB:luo_nqe_2022}.
% Another challenge is the attributes of entities and relations which are largely undiscussed, the attributes can be categorical~\citep{hu_type-aware_2022} or numerical~\citep{DBLP:conf/kdd/BaiLLYYS23}.
% These features are largely unexplored, leading to the consequence that current query expressiveness is strictly limited compared to database-level query languages . 
\textbf{(ii) Support of functions}
The corresponding research gap is the support for functions in the query -- a function can output nodes, numbers, semantic units, or data of more advanced modality -- for example, the \texttt{AVG} and \texttt{COUNT} functions in SQL but not in current CQA models. We have noted one preliminary research trying to fill this gap~\cite{NGDB:DBLP:conf/pkdd/DemirWLNH23}.

\subsection{Minimal Assumption for Broad Generalization}

Previous case studies showcase the close entanglement of the neural modules and the query types they support syntactically. In other words, the key to generalization is minimizing the query families' assumptions and the inductive biases of the neural part of NGDB. We present two types of methods with minimal assumptions.

\textbf{Neuro-symbolic Methods.} NGDB implies that the underlying database is a graph, meaning neural modules solely modeling the graph itself impose no assumptions on the query family it might support. Such neural modules include link predictors or knowledge graph embeddings that map a triple $(s, p, o)$ of subject, predicate, and object into a score~\cite{NGDB:wang2017knowledge}. Therefore, the critical design task of NGDB with such modules is revising the algorithms into the neuro-symbolic forms with the scores produced by link predictors~\cite{NGDB:bai_answering_2023,NGDB:yin_rethinking_2023}. An apparent and more decomposed approach is to derive an instance of a classic graph database using the link predictor, and all previous research in graph databases applies directly. Notably, the neuro-symbolic approach achieves the same level of generalizability in queries as the classic database research.

\textbf{Sequence Models.} Language models or sequence models are general-purpose models and thus further disentangle the inductive bias of neural modules and the specific task (queries in NGDB). Such models support the sequence inputs, which cover inputs from all possible kinds of query types. However, the performance on specific query types is transferred from designing neural architectures to curating the training datasets. The cost is transferred from the complex inference algorithms to the training phase~\cite{NGDB:bai2023sequential}.

\subsection{Learning Aspects of Generalization}
From the machine learning perspective, one new issue is uniformly improving the performance of all queries of a particular query family under the analogy of query types as tasks. The approach towards this goal also varies for different methods. {For neuro-symbolic approaches}, the generalization will be improved coherently as link prediction performance improves. {For neural methods}, the challenge of generalization is the same as multi-task learning. Query embeddings, as a particular case of neural methods, recent works propose adopting set operators with meta-learning, yielding the solution of meta operators~\cite{NGDB:yin2024meta}.

\vspace{-.5em}
\section{Challenge 4: Privacy and Security}
\vspace{-.5em}

\subsection{Database Privacy}
Privacy in data storage refers to protecting sensitive information from unauthorized access and misuse~\cite{NGDB:olivier2002database}. Traditional databases are facing several privacy risks, which can be categorized into:
{(1) Unauthorized Access}~\cite{NGDB:bertino2011access}: Unauthorized access to databases can result in large-scale data leakage, exposing sensitive personal information. 
{(2) Insider Threats}~\cite{NGDB:senator2013detecting}: Employees with legitimate access may misuse their privileges, either intentionally or unintentionally compromising data privacy. 
{(3) Data Inference Attacks}~\cite{NGDB:naveed2015inference}: Attackers can employ various techniques to deduce sensitive information from seemingly innocuous data.

To mitigate privacy risks, several protection methods have been developed: {(1) Data Anonymization}~\cite{NGDB:murthy2019comparative}: Techniques such as k-anonymity~\cite{NGDB:sweeney2002k} and l-diversity~\cite{NGDB:machanavajjhala2007diversity} help mask individual identities within datasets, making it harder to trace data back to specific individuals.
{(2) Encryption}~\cite{NGDB:basharat2012database}: Data encryption ensures that unauthorized parties cannot access sensitive information even if they breach a database.
{(3) Access Control}~\cite{NGDB:bertino2011access}: Access control restricts data access to authorized users only, reducing the risk of insider threats.
{(4) Differential Privacy}~\cite{NGDB:dwork2006differential}: This approach adds noise to data outputs, ensuring that the presence or absence of an individual in a dataset does not significantly affect the results of queries.

\subsection{New Privacy Challenges in NGDBs}
Graph databases, while offering advantages in managing complex relationships, introduce specific privacy risks: {(1) Link Prediction Attacks}~\cite{NGDB:lin2020adversarial}: Adversaries can use machine learning models to predict hidden relationships within the graph, potentially uncovering private connections.
{(2) Structural Attacks}~\cite{NGDB:zhao2021structural}: Even when the data content is anonymized, the graph’s structure itself can reveal sensitive insights.
The unique structure of graph data amplifies these risks, as the relationships between entities can reveal information that is not immediately apparent from isolated data points.
Neural Graph Databases (NGDBs) represent a significant advancement in data management, combining the strengths of traditional graph databases with the capabilities of neural networks. The exploration of privacy issues in NGDBs remains largely underdeveloped, with significant gaps in research addressing potential vulnerabilities and mitigation strategies.

\textbf{Potential Attacks.} One of the primary strengths of NGDBs is their ability to generalize from incomplete data by inferring hidden relationships. While this capability can enhance data retrieval and knowledge discovery, it also poses significant privacy risks~\cite{NGDB:hu2024privacy}:
{(1) Model Inversion Attacks}~\cite{NGDB:fang2024privacy}: Neural models can be susceptible to inversion attacks, where an adversary uses access to the model to recover the graph data used for NGDB training.
{(2) Membership Inference Attacks}~\cite{NGDB:hu2022membership}: Attackers may infer whether a particular data point (node or edge) was part of the training data, revealing sensitive information in NGDBs.
{(3) Embedding Leakage}~\cite{NGDB:song2020information}: The embeddings generated by NGDBs to represent nodes and relationships can leak sensitive information, as these embeddings often capture detailed structural and content-based features of the graph stored.

\textbf{Promising Defenses.}
{(1) Differential Privacy in NGDBs}: Extending differential privacy techniques to protect neural graph databases is a key research direction. Adding noise to the model parameters or gradients during training can help mitigate membership inference and model inversion attacks~\cite{NGDB:truex2019effects, NGDB:wang2015regression}.
{(2) Embedding Obfuscation}: Techniques to obfuscate embeddings without losing their utility for answering complex queries need to be developed to prevent leakage of sensitive information~\cite{NGDB:hu2023independent}.
{(3) Private Distribute Training}: Privacy problems in distributed NGDBs need further development~\cite{NGDB:hu2024fedcqa}. Federated learning, including Secure Multi-Party Computation (SMPC)~\cite{NGDB:goldreich1998secure} and Homomorphic Encryption (HE)~\cite{NGDB:acar2018survey} techniques, can be adapted to NGDBs to ensure that data is processed without being revealed.

\textbf{Evaluation Benchmarks.}
Another significant challenge in NGDBs is the evaluation of privacy protection efficacy. Assessing the effectiveness of privacy-preserving mechanisms requires robust benchmarks that can accurately measure both privacy protection and the quality of retrieved data. However, such benchmarks are currently lacking in the field. 
To address this challenge, standardized evaluation metrics and datasets should be developed that can facilitate comprehensive testing of privacy-preserving techniques in NGDBs. Establishing reliable benchmarks will provide insights into the strengths and weaknesses of different approaches, ultimately guiding future developments in privacy protection.
\vspace{-.5em}
\section{Challenge 5: Scaling for Higher Complexity}
\vspace{-.5em}

% Challenge scalability

% 1. training cost
%     total cost of train = \# training set * (forward + backward)
%     how large is the training set?
% 2. inference cost
%     Data complexity and query complexity -> large graph challenge and complexity of query family.
%     1. state complexity class of (tree-form (1-answer), EFO (EFO-1 + efok) (SQL, SPARQL, ...)
    
% search, classic results ...
% sequence model, data complexity (token dimension at most linear) query complexity (linear)

% Methodology types
% 1. size of models
% 2. size of datasets
% 3. size of cost
% 4. performance

In deep learning, neural scaling law is an empirical law that describes the performance of neural models improves with the number of parameters, training dataset size, and training cost~\cite{NGDB:hestness2017deep,NGDB:kaplan2020scaling}. During the development of the NGDB model, scaling is also a major thread, primarily encompassing the scaling of parameter number, query data size, and training costs. The query embedding methods and sequence models often scale the training costs in the training stage, including the model parameters and queries. In contrast, the neuro-symbolic methods often scale the computation cost over the test stage to improve the performance.  We mainly discuss how to scale these models further, particularly when the query structure becomes increasingly complex~\cite{NGDB:yin_rethinking_2023, NGDB:yin_textefo_k-cqa_2023} and the magnitude of the knowledge databases becomes very large~\cite{NGDB:ren_smore_2022}. Specifically, we introduce the complexity of these models in the training and inference stages and discuss their efficiency and scalability challenges.

\textbf{Data Scaling in the Training Stage.}
Both query embedding and sequence models are trained from scratch, requiring many sampled queries as training data. The quality and size of these training queries are crucial, and they typically encompass various query types. The NGDB models generally use the same dataset, with the basic 1p query type enumerating the entire knowledge graph~\cite{NGDB:ren_query2box_2020}. To incorporate new features such as negation~\cite{NGDB:ren_beta_2020}, cyclic queries~\cite{NGDB:yin_rethinking_2023}, and multivariable queries~\cite{NGDB:yin_textefo_k-cqa_2023}, it is essential to sample query types that include these features. Materializing training queries becomes infeasible as the knowledge graph grows, and sampling logical queries is incompatible with traditional single-hop frameworks based on graph partitioning. To address this challenge, SMORE~\cite{NGDB:ren_smore_2022} proposes a scalable framework that efficiently samples training data on the fly with high throughput.
In contrast, neuro-symbolic methods primarily rely on pre-training for the knowledge graph completion task and depend on search algorithms to address general logical queries.

\textbf{Test Time Scaling in Inference Stage.}
We first introduce the notion of query complexity and data complexity~\cite{NGDB:abiteboul1995foundations}. Data complexity captures the relation between the time complexity and the database size $|E|$ (number of the edges) when the query is fixed. In contrast, query complexity is assessed based on the size of the query $|Q|$ (number of the predicates) when assuming the database is fixed. When discussing the complexity, the query is restricted to tree-formed queries and EFO-1 queries that we have discussed before.
The complexity of neural symbolic search is well studied. The complexity for tree-formed queries is $O(|Q||E|)$. Such approaches~\cite{NGDB:yin_rethinking_2023,NGDB:bai_answering_2023} require $O(|Q|)$ search steps, while each step requires a search over the database, which is $O(|E|)$. For the general EFO-1 query, the cyclic query makes the general complexity particularly hard and results in $O(|E|^{|Q|})$ time, which is polynomial in data but exponential in query. One distinct feature of query embeddings and sequence models compared to neuro-symbolic methods is the disentanglement of the $|E|$ term and $|Q|$ term. Notably, the neural network encoder~\cite{NGDB:wang2023logical} or sequence model~\cite{NGDB:bai2023sequential} work on the query directly, which is usually $O(|Q|)$ to encode query information and $O(|E|)$ to decode the answer by embedding comparison. However, this great advantage in inference time complexity of query embeddings and neural symbolic models comes from the additional and usually resource-consuming training procedures.

\vspace{-.5em}
\section{Challenge 6: Distributed NGDB System}
\vspace{-.5em}
\subsection{Scenario Features and System Requirements}

\paragraph{Scenario Features.} NGDB is targeted at a scenario where users can simultaneously conduct graph data management and graph inference. We identify four features of such a scenario that significantly affect the system design. {(1) Hybrid symbolic and neural operation}~\cite{NGDB:ren2023neural}. Users can input queries requiring algebraic, neural, or hybrid computation; {(2) Massive graph data and embeddings}. Not only do the graph data of different domain knowledge exhibit tremendous scale~\cite{NGDB:bollacker2008freebase}, but also various types of embeddings~\cite{NGDB:khan2023knowledge} of these graph data further enlarge the volume; {(3) Read intensive workload}. During the serving stage, most of the graph data and embeddings are queried more frequently rather than updated~\cite{NGDB:tian2023world}; {(4) Dynamic workload fluctuation}. Different parts of the graph data and embeddings are accessed in different time slots and the number of online users and frequency and data volume of one query fluctuate~\cite{NGDB:guo2022manu}. 

\paragraph{System Requirements.} The neural graph database system should fulfill the following requirements to handle these features effectively and efficiently. {(1) Co-located graph and embedding management.} The NGDB system should support symbolic graph data and neural embedding management. {(2) High query performance.} The latency of a single query and system throughput for numerous tenants serving massive data should be optimized. {(3) Scalability.}. The hardware resource management should be scalable to handle workload fluctuation, especially computational resources, cost-efficiently. Challenges are introduced to the system design of neural graph databases to implement these system features.

\subsection{Challenges of System Design}

\paragraph{User Interface Design.} Existing vector databases provide SQL-like interfaces and parameterized API\cite{NGDB:pan2024survey}, while most of the interfaces mainly focus on relational data. Graph databases provide numerous interfaces\cite{NGDB:khan2023knowledge}, but there is little experience in combining neural operations into symbolic graph operations. It is essential to design highly expressive declarative user interfaces as well as programming interfaces.

\paragraph{Query-Oriented Distributed Storage.} Due to the massive volume of graph data and corresponding embeddings, which is out of the capacity of standalone storage, distributed storage is an indispensable mechanism of NGDB. Under read-intensive workload, partitioning  (or sharding), acting as a distributed index, tailored for most frequent and costly types of query could remarkably reduce the intermediate data transfer, consequently enhancing the overall latency and throughput\cite{NGDB:pan2024survey}. Practices in graph database community\cite{NGDB:ammar2016graph, NGDB:fu2019geabase, NGDB:li2022bytegraph, NGDB:sun2023design} and vector database community concludes valuable principles and strategies on distributed storage and indexing of graph data and embedding separately. However, the hybrid storage of both data types is not explored, especially in circumstances where hybrid queries, requiring both symbolic and neural processes, are of evident importance. A typical example question is about whether embeddings and raw graph data shall be co-located. Although some open source graph database\cite{NGDB:neo4j, NGDB:tigergraph} and vector databases\cite{NGDB:wang2021milvus, NGDB:guo2022manu, NGDB:qdrant, NGDB:weaviate} could be utilized as standalone storage engine in NGDB, partitioning should be carefully designed under specific query workload.

\paragraph{Distributed Graph Computing and Inference.} There are abundant works on distributed graph query and analysis with various algorithms in the graph database community\cite{NGDB:bouhenni2021survey, NGDB:meng2024survey}. However, distributed system support for knowledge graph inference has not been adequately explored. Atom\cite{NGDB:zhou2024atom} points out a key observation that query embedding is the performance bottleneck, which shall be one of the considerations in NGDB query execution. On the base of these two kinds of computation optimization, when encountered with hybrid queries requiring both computation, query planning, and scheduling for maximized parallelism and minimized network communication overhead, still remain an unexplored direction.
There are some preliminary practice cases in relational databases~\cite{NGDB:yuan2024nsdb, NGDB:zhao2024neurdb, NGDB:liu2024optimizing}, which consider the optimization with neural operators but are still far from mature. 

\paragraph{Elastic Scalability.} To deal with dynamic workload fluctuation, fine-grained elasticity is of great importance to distributed NGDB systems, in which case on-demand resource provision helps reduce the cost\cite{NGDB:zhang2023cost} of NGDB service. Besides, not all the massive data are simultaneously accessed. There are evident biases and data heat shifts in database serving scenarios. Therefore, we argue that being cloud-native with elastic scalability is a crucial requirement for the NGDB system. Manu\cite{NGDB:guo2022manu} detects such workload fluctuation in industrial applications, thus fully embracing the mechanisms of elastic scalability via dedicated abstraction of hardware resource management, including GPU, CPU, and disk. Besides, storage-computation-separation is essential for cloud-native databases\cite{NGDB:li2019cloud}. It is essential to explore the combination of these separate practices. Additionally, the trade-off between latency and elasticity is a critical concern since practices in vector databases reveal that embedding management requires a large memory occupation.

\vspace{-.5em}
\section{Challenge 7: Compatibility of NGDB with Traditional Graph Database}
\vspace{-.5em}

% \subsection{Structural and Operational Compatibility}
Like graph databases, Neural Graph Databases (NGDB) are another way of the data model that derives the properties from the existing graphs, including nodes and edges, to represent entities and their relationships~\cite{NGDB:ren2023neural}. This structural consistency makes migrating and interoperating data between the two databases relatively easy. In terms of interfaces, NGDB can maintain support for standard graph query languages~\cite{NGDB:neo4j, NGDB:10.1145/3183713.3190657} while providing vectorized query capabilities, allowing users to query and operate in familiar languages. In terms of operations, traditional CRUD operations remain fully functional, with the reasoning function of neural networks serving as enhanced features. For instance, conventional graph databases provide foundational support in query processing through mature storage and indexing technologies, while NGDB handles queries requiring missing link inference. Such compatibility design will enable a seamless system transition,  where users can migrate to get NGDB capabilities without completely reconstructing existing applications.
However, NGDB faces several challenges with traditional graph databases:

\textbf{Novel Query Interface.} Incorporating deep learning and graph neural networks extends beyond conventional graph database functionalities, requiring novel interfaces for deep learning-based queries and inference. This creates compatibility issues when attempting to reuse existing query languages, highlighting the need to develop new query languages or extend current ones~\cite{NGDB:DBLP:journals/corr/abs-1909-01315,NGDB:10.1145/3183713.3190657}.

\textbf{Performance-Consistency Trade-off.} While traditional graph databases are optimized for storage and querying~\cite{NGDB:10.1145/3183713.3190654}, they may struggle to meet performance requirements when handling large-scale graph-based deep-learning tasks. NGDB emphasizes representation learning on nodes and edges~\cite{NGDB:ren2023neural}, requiring consideration of high-performance computing and distributed training paradigms. For instance, during conventional CRUD operations, NGDB may need to update node and relation embeddings, introducing additional computational overhead. Moreover, integrating neural components introduces temporal consistency challenges, where model updates may lead to temporary discrepancies between the base graph data and learned representations. Finding an optimal balance between consistency guarantees and computational efficiency remains a considerable challenge for NGDB systems.

\section{Challenge 8: Grounding to Vectors with NGDB}

Grounding natural language to knowledge bases has been extensively studied in conventional graph databases. Traditional approaches typically handle different grounding scenarios: hypothesis or query grounding (with free variables) \cite{NGDB:zhong2019reasoning, NGDB:ren_query2box_2020}, and entity \cite{NGDB:shen2014entity} or event \cite{NGDB:bai2023complexqueryansweringeventuality, NGDB:jiayang2024eventground}).
With the emergence of NGDBs, where structural information and semantic content are encoded as vectors, the grounding process faces new challenges and opportunities. Recent work \cite{NGDB:ren2023neural} introduces a neural graph engine that learns query planning and execution strategies through interactions with Neural Graph Storage.
However, grounding to general NGDBs still presents several unique challenges. 

\textbf{Semantic Granularity and Disambiguation.}
Semantic granularity and disambiguation pose fundamental difficulties. 
The grounding process must accurately translate natural language queries into appropriate vector representations while determining suitable levels of semantic granularity, such events, propositions, etc. \cite{NGDB:chen2023dense, NGDB:jiayang2024eventground}. 
This challenge is compounded by the need to handle abstraction and polysemy when mapping linguistic elements to vector spaces, as meanings can vary significantly based on context.

\textbf{Compositional Semantics and Reasoning.}
Second, compositional semantics and reasoning path selection present significant challenges. 
NGDBs must effectively represent complex multi-hop relations while maintaining transitivity and logical consistency in vector operations. 
The system needs to identify relevant paths in the vector space for query resolution, which becomes particularly challenging when dealing with multiple possible reasoning paths. 
In addition, determining appropriate termination criteria for path exploration is crucial for both efficiency and accuracy.

\textbf{Interpretation and Groundedness Evaluation.}
The third challenge is around interpretation and groundedness evaluation. 
The system is expected to reliably convert vector-based results back to natural language while providing clear explanations for its reasoning process. 
Additionally, it needs to report the level of groundedness for each grounding operation, ensuring semantic fidelity is maintained throughout the process. 
This is particularly important for applications requiring high precision and explainability.

\vspace{-.5em}
\section{Challenge 9: Adapting NGDB to LLM}
\vspace{-.5em}

This section explores the integration of Neural Graph Databases (NGDBs) with Large Language Models (LLMs) to enable joint reasoning and Retrieval-Augmented Generation (RAG). NGDBs can serve as retrieval modules for LLMs, leveraging structured data and reasoning capabilities to enhance generated outputs' accuracy, scalability, and contextual relevance. Joint learning of LLMs and NGDBs involves training these systems within a unified framework to combine natural language understanding with advanced logical reasoning.

\textbf{NGDB-RAG: Definition and Components.}
NGDB-RAG (Neural Graph Database - Retrieval-Augmented Generation) is a system that integrates NGDBs with LLMs to enhance both retrieval and generation tasks. The NGDB-RAG system is composed of three main components. The first is the neural graph storage, which stores embeddings of nodes and edges in the graph. These embeddings capture both local and global structural relationships within the graph, providing a rich representation of the data. The second component is the neural query engine, which tries formulating and processing logical queries in the embedding space. This engine enables flexible modeling and supports logical operations such as conjunction, disjunction, and negation, allowing for robust retrieval even in incomplete or noisy graphs. The third component is integrating with LLMs, where NGDB reasoning results are incorporated into the language model. This integration can be achieved through text-based methods, by converting structured data into natural language, or through vector-based methods, by embedding structured data as vectors for direct input into the LLM.

\textbf{Functionality of NGDB-RAG.}
NGDB-RAG enhances retrieval by utilizing the structured relationships in NGDBs to perform advanced reasoning tasks. Unlike traditional RAG systems that rely on document similarity, NGDB-RAG leverages the intricate dependencies within knowledge graphs to retrieve more accurate and contextually relevant information. In the generation process, NGDB-RAG integrates structured knowledge and reasoning capabilities from NGDBs to improve the generated text's factual accuracy and logical consistency while reducing hallucinations. Furthermore, NGDB-RAG is designed to handle large-scale graphs and supports various query types, including temporal, spatial, and numerical reasoning, ensuring scalability and expressiveness in practical applications.

\textbf{Joint Learning Framework.}
The joint learning framework of NGDBs and LLMs employs a co-training approach where both systems share parameters or representation spaces to enable collaborative learning. Improvements in one component positively influence the other, creating a feedback loop that enhances the overall system. The combined training objective is expressed as: $L_{\text{total}} = L_{\text{LLM}} + \lambda L_{\text{NGDB}}$.
In this equation, \( L_{\text{LLM}} \) represents the loss associated with the language model, typically the cross-entropy loss for next-token prediction. \( L_{\text{NGDB}} \) denotes the loss related to NGDB reasoning tasks, such as the error between predicted and true query answers. The hyperparameter \( \lambda \) controls the balance between the two loss components. The objective of this joint training is to improve the reasoning capabilities of the NGDB while enhancing the LLM performance.

Future work aims to develop the co-training framework further to enable simultaneous training of NGDB reasoning engines and LLMs, ensuring parameter sharing and collaborative learning. Efforts are also being made to refine the combined loss function to balance language modeling and reasoning tasks better, enhancing both components' performance. Integration modules are being developed to incorporate NGDB reasoning results into LLMs through text-based and vector-based methods. These advancements are expected to create a unified system capable of performing advanced reasoning and generating high-quality, contextually accurate text.

% \subsubsection{Progress in NGDB-RAG Development}

% Significant advancements have been made in enhancing NGDB reasoning engines, including scalable query encoding and improved expressiveness to handle diverse and complex queries. Additionally, extensive benchmark datasets and evaluation metrics have been developed to assess the performance of NGDBs in complex query answering and their integration with LLMs.

% Current and future efforts focus on developing and evaluating integration modules seamlessly incorporating NGDB reasoning results into LLMs through text-based and vector-based knowledge injection methods. A co-training framework is being established to jointly train NGDBs and LLMs, optimizing the combined system for enhanced reasoning and generation capabilities. Furthermore, NGDB-RAG systems are applied to various real-world domains, including general knowledge, medical, and financial sectors, to ensure scalability and domain-specific effectiveness.

\section{Challenge 10: Smart Neural Graph Databases}

% \subsection{Applications of Agentic NGDB}
Benefited from its rich functionalities, Agentic NGDB offers a wide range of applications across  domains:

\indent $\bullet$ \textbf{Autonomous Data Management}: Agentic NGDB can autonomously manage complex datasets, optimize query execution, and organize storage structures without human intervention. This is particularly useful in large-scale systems where manual optimization is impractical.

\indent $\bullet$ \textbf{Personalized Recommendations}: Through continuous learning, Agentic NGDB can provide real-time personalized recommendations by analyzing user preferences and graph-based relationships. This is crucial in e-commerce and social networks, where tailored experiences drive user engagement \cite{NGDB:bai2024understandingintersessionintentionscomplex}.

\indent $\bullet$ \textbf{Complex Event Processing}: Agentic NGDB is well-suited for handling complex event processing \cite{NGDB:bai2023complexqueryansweringeventuality}, where multiple events and data streams need to be analyzed in real-time. By leveraging their semantic understanding and neural inference, Agentic NGDB can identify correlations and patterns across seemingly unrelated events, making them valuable in cybersecurity, fraud detection, and IoT systems.

\vspace{-.5em}
\section{Conclusion}
\vspace{-.5em}

Agentic Neural Graph Databases (Agentic NGDBs) represent an advancement in data management, building on traditional graph databases and Neural Graph Databases (NGDBs) by introducing autonomy, continuous learning, and advanced reasoning.

This paper identifies ten key challenges to realizing Agentic NGDBs, including semantic representation, abductive reasoning, generalization across query types, scalability, privacy, and integration with foundation models like large language models (LLMs). Ensuring compatibility with traditional databases, grounding knowledge in vectors, and developing distributed systems are essential for achieving robust and scalable solutions.

By overcoming these challenges, Agentic NGDBs can transform modern data-driven applications. Their ability to autonomously generate and execute queries, support continuous learning, and integrate symbolic and neural reasoning offers new possibilities in autonomous data management, personalized recommendations, and complex event processing. These advancements promise to redefine how we manage, query, and reason over interconnected data for the future.

\bibliographystyle{unsrt}
\bibliography{ref.bib}

\end{document}